\documentclass[10pt,twocolumn,letterpaper]{article}

\usepackage{cvpr}
\usepackage{times}
\usepackage{epsfig}
\usepackage{graphicx}
\usepackage{amsmath}
\usepackage{amssymb}
\usepackage{color}
\usepackage{booktabs} 
\usepackage{multirow}
\usepackage{array}
\usepackage{verbatim}
\usepackage{caption}
\usepackage{array, siunitx, boldline, hhline}
\usepackage{threeparttable}
\usepackage[pagebackref=true,breaklinks=true,letterpaper=true,colorlinks,bookmarks=false]{hyperref}
\usepackage{authblk}
\usepackage[english]{babel}
\usepackage{blindtext}

\usepackage[breaklinks=true,bookmarks=false]{hyperref}

\cvprfinalcopy 

\def\cvprPaperID{****} 

\newcommand\blfootnote[1]{%
  \begingroup
  \renewcommand\thefootnote{}\footnote{#1}%
  \addtocounter{footnote}{-1}%
  \endgroup
}
\begin{document}

\title{MirrorGAN: Learning Text-to-image Generation by Redescription}

\author[1,3]{\normalsize{Tingting~Qiao}}
\author[2,3,*]{\normalsize{Jing~Zhang}}
\author[1,*]{\normalsize{Duanqing~Xu}}
\author[3]{\normalsize{Dacheng~Tao}}
\affil[1]{\normalsize{College of Computer Science and Technology, Zhejiang University, China}}
\affil[2]{\normalsize{The institute of System Science and Control Engineering, Hangzhou Dianzi University, China}}
\affil[3]{UBTECH Sydney AI Centre, School of Computer Science, FEIT, University of Sydney, Australia}
\affil[ ]{\small \texttt {qiaott@zju.edu.cn, jing.zhang@uts.edu.au, xdq@zju.edu.cn, dacheng.tao@sydney.edu.au}}
\maketitle

\begin{abstract}
Generating an image from a given text description has two goals: visual realism and semantic consistency. Although significant progress has been made in generating high-quality and visually realistic images using generative adversarial networks, guaranteeing semantic consistency between the text description and visual content remains very challenging. In this paper, we address this problem by proposing a novel global-local attentive and semantic-preserving text-to-image-to-text framework called MirrorGAN. MirrorGAN exploits the idea of learning text-to-image generation by redescription and consists of three modules: a semantic text embedding module (STEM), a global-local collaborative attentive module for cascaded image generation (GLAM), and a semantic text regeneration and alignment module (STREAM). STEM generates word- and sentence-level embeddings. GLAM has a cascaded architecture for generating target images from coarse to fine scales, leveraging both local word attention and global sentence attention to progressively enhance the diversity and semantic consistency of the generated images. STREAM seeks to regenerate the text description from the generated image, which semantically aligns with the given text description. Thorough experiments on two public benchmark datasets demonstrate the superiority of MirrorGAN over other representative state-of-the-art methods. 
\end{abstract}
\blfootnote{1.The work was performed when Tingting Qiao was a visiting student at UBTECH Sydney AI Centre in the School of Computer Science, FEIT, in the University of Sydney}
\blfootnote{2.*corresponding author}
\section{Introduction}
\begin{figure}[tb!]
\centering
\noindent\includegraphics[width=0.9\columnwidth]{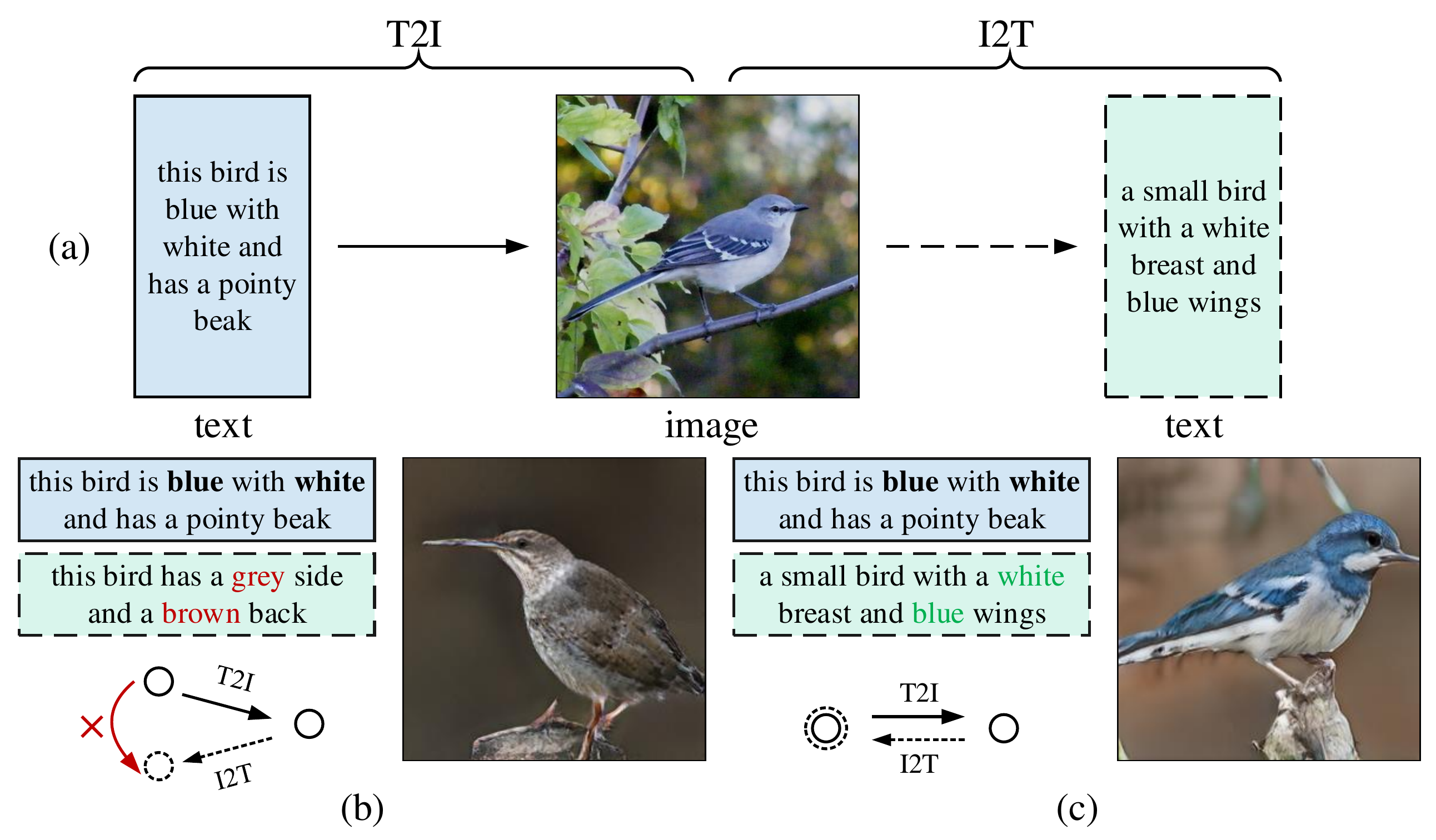}
\protect\caption{(a) Illustration of the mirror structure that embodies the idea of learning text-to-image generation by redescription. (b)-(c) Semantically inconsistent and consistent images/redescriptions generated by \cite{xu2017attngan} and the proposed MirrorGAN, respectively.}
\label{fig:Motivation}
\end{figure}

Text-to-image (T2I) generation refers to generating a visually realistic image that matches a given text description. Due to its significant potential in a number of applications but its challenging nature, T2I generation has become an active research area in both natural language processing and computer vision communities. Although significant progress has been made in generating visually realistic images using generative adversarial networks (GANs) such as in \cite{zhang2017stackgan,zhang2018photographic,xu2017attngan,hong2018inferring}, guaranteeing semantic alignment of the generated image with the input text remains challenging.

In contrast to fundamental image generation problems, T2I generation is conditioned on text descriptions rather than starting with noise alone. Leveraging the power of GANs \cite{goodfellow2014generative}, different T2I methods have been proposed to generate visually realistic and text-relevant images. For instance, Reed \etal proposed to tackle text to image synthesis problem by finding a visually discriminative representation for the text descriptions and using this representation to generate realistic images \cite{reed2016generative}. Zhang \etal proposed StackGAN to generate images in two separate stages \cite{zhang2017stackgan}. Hong \etal proposed extracting a semantic layout from the input text and then converting it into the image generator to guide the generative process \cite{hong2018inferring}. Zhang \etal proposed training a T2I generator with hierarchically nested adversarial objectives \cite{zhang2018photographic}. These methods all utilize a discriminator to distinguish between the generated image and corresponding text pair and the ground truth image and corresponding text pair. However, due to the domain gap between text and images, it is difficult and inefficient to model the underlying semantic consistency within each pair when relying on such a discriminator alone. Recently, the attention mechanism \cite{xu2017attngan} has been exploited to address this problem, which guides the generator to focus on different words when generating different image regions. However, using word-level attention alone does not ensure global semantic consistency due to the diversity between text and image modalities. Figure~\ref{fig:Motivation} (b) shows an example generated by \cite{xu2017attngan}.

T2I generation can be regarded as the inverse problem of image captioning (or image-to-text generation, I2T) \cite{xu2015show,vinyals2015show,karpathy2015deep}, which generates a text description given an image. Considering that tackling each task requires modeling and aligning the underlying semantics in both domains, it is natural and reasonable to model both tasks in a unified framework to leverage the underlying dual regulations. As shown in Figure~\ref{fig:Motivation} (a) and (c), if an image generated by T2I is semantically consistent with the given text description, its redescription by I2T should have exactly the same semantics with the given text description. In other words, the generated image should act like a mirror that precisely reflects the underlying text semantics. Motivated by this observation, we propose a novel text-to-image-to-text framework called MirrorGAN to improve T2I generation, which exploits the idea of learning T2I generation by redescription.

MirrorGAN has three modules: STEM, GLAM and STREAM. STEM generates word- and sentence-level embeddings, which are then used by the GLAM. GLAM is a cascaded architecture that generates target images from coarse to fine scales, leveraging both local word attention and global sentence attention to progressively enhance the diversity and semantic consistency of the generated images. STREAM tries to regenerate the text description from the generated image, which semantically aligns with the given text description. 

To train the model end-to-end, we use two adversarial losses: visual realism adversarial loss and text-image paired semantic consistency adversarial loss. In addition, to leverage the dual regulation of T2I and I2T, we further employ a text-semantics reconstruction loss based on cross-entropy (CE). Thorough experiments on two public benchmark datasets demonstrate the superiority of MirrorGAN over other representative state-of-the-art methods with respect to both visual realism and semantic consistency.

The contributions of this work can be summarized as follows:

$\bullet$ We propose a novel unified framework called MirrorGAN for modeling T2I and I2T together, specifically targeting T2I generation by embodying the idea of learning T2I generation by redescription.

$\bullet$ We propose a global-local collaborative attention model that is seamlessly embedded in the cascaded generators to preserve cross-domain semantic consistency and to smoothen the generative process.

$\bullet$ Except commonly used GAN losses, we additionally propose a CE-based text-semantics reconstruction loss to supervise the generator to generate visually realistic and semantically consistent images. Consequently, we achieve new state-of-the-art performance on two benchmark datasets.

\begin{figure*}[tb!]
\centering
\noindent\includegraphics[width=1.9\columnwidth]{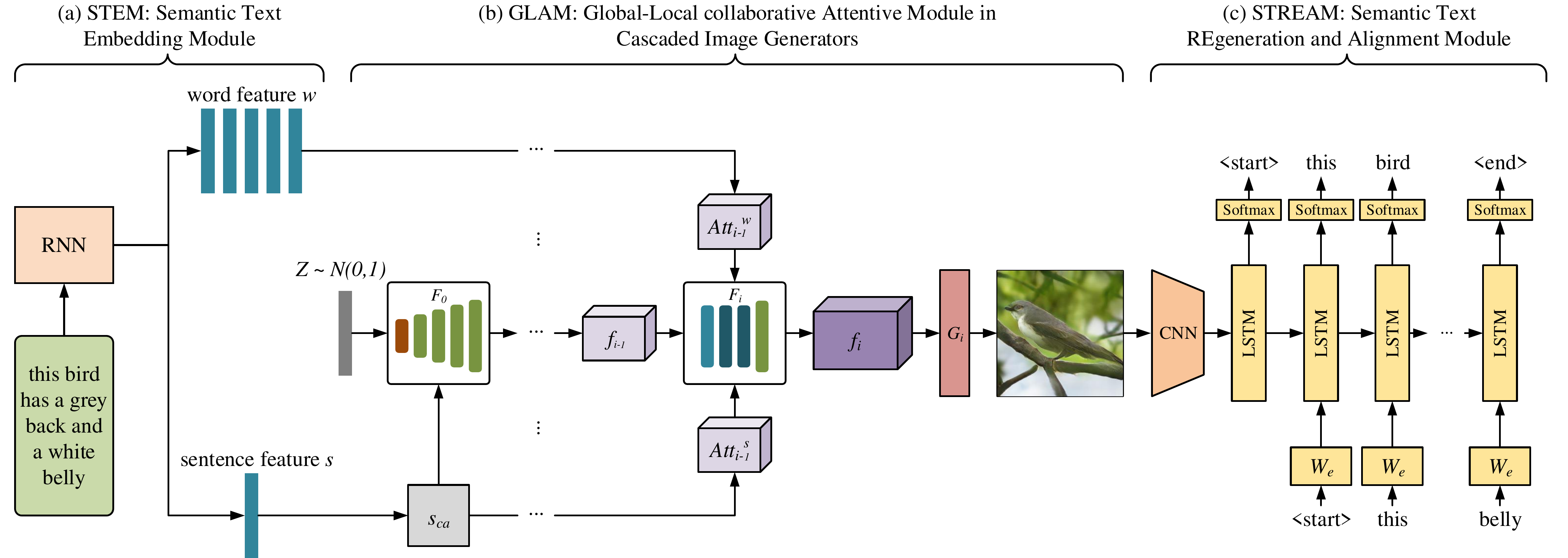}
\protect\caption{Schematic of the proposed MirrorGAN for text-to-image generation.}
\label{fig:Model}
\end{figure*}

\section{Related work}
Similar ideas to our own have recently been used in CycleGAN and DualGAN, which handle the bi-directional translations within two domains together \cite{zhu2017unpaired,yi2017dualgan,Almahairi2018AugmentedCL}, significantly advance image-to-image translation \cite{isola2016image,taigman2016unsupervised,johnson2016perceptual,Yu2018Inpainting,qiao2019ancient}. Our MirrorGAN is partly inspired by CycleGAN but has two main differences: 1) we specifically tackle the T2I problem rather than image-to-image translation. The cross-media domain gap between text and images is probably much larger than the one between images with different attributes, $e.g.,$ styles. Moreover, the diverse semantics present in each domain make it much more challenging to maintain cross-domain semantic consistency. 2) MirrorGAN embodies a mirror structure rather than the cycle structure used in CycleGAN. MirrorGAN conducts supervised learning by using paired text-image data rather than training from unpaired image-image data. Moreover, to embody the idea of learning T2I generation by redescription, we use a CE-based reconstruction loss to regularize the semantic consistency of the redescribed text, which is different from the L1 cycle consistency loss in CycleGAN, which addresses visual similarities.

Attention models have been extensively exploited in computer vision and natural language processing, for instance in object detection \cite{oliva2003top,deubel1996saccade,Liu2018DecideNet,Zhang2018ObjectDetection}, image/video captioning \cite{xu2015show,gao2017video,Wang2018Captioning}, visual question answering \cite{anderson2017bottom,xu2016ask,yang2016stacked,qiao2018exploring}, and neural machine translation \cite{luong2015effective,firat2016multi}. Attention can be modeled spatially in images or temporally in language, or even both in video- or image-text-related tasks. Different attention models have been proposed for image captioning to enhance the embedded text feature representations during both encoding and decoding. Recently, Xu \etal proposed an attention model to guide the generator to focus on different words when generating different image subregions \cite{xu2017attngan}. However, using only word-level attention does not ensure global semantic consistency due to the diverse nature of both the text and image modalities, $e.g.,$ each image has 10 captions in CUB and 5 captions in COCO,however, they express the same underlying semantic information. In particular, for multi-stage generators, it is crucial to make ``semantically smooth'' generations. Therefore, global sentence-level attention should also be considered in each stage such that it progressively and smoothly drives the generators towards semantically well-aligned targets. To this end, we propose a global-local collaborative attentive module to leverage both local word attention and global sentence attention and to enhance the diversity and semantic consistency of the generated images. 

\section{MirrorGAN for text-to-image generation}
As shown in Figure~\ref{fig:Model}, MirrorGAN embodies a mirror structure by integrating both T2I and I2T. It exploits the idea of learning T2I generation by redescription. After an image is generated, MirrorGAN regenerates its description, which aligns its underlying semantics with the given text description. Technically, MirrorGAN consists of three modules: STEM, GLAM and STREAM. Details of the model will be introduced below.

\subsection{STEM: Semantic Text Embedding Module}
First, we introduce the semantic text embedding module to embed the given text description into local word-level features and global sentence-level features. As shown in the leftmost part of Figure~\ref{fig:Model}, a recurrent neural network (RNN) \cite{cho2014learning} is used to extract semantic embeddings from the given text description $T$, which include a word embedding $w$ and a sentence embedding $s$. 
\begin{equation}
w,s = RNN\left( T \right),
\label{eq:textEmbedding}
\end{equation}
where $T = \left\{ {{T_l}\left| {l = 0, \ldots ,L - 1} \right.} \right\}$, $L$ represents the sentence length, $w = \left\{ {{w^l}\left| {l = 0, \ldots ,L - 1} \right.} \right\} \in {\mathbb{R}^{D \times L}}$ is the concatenation of hidden state ${{w^l}}$ of each word, $s\in \mathbb{R}^{D}$ is the last hidden state, and $D$ is the dimension of ${{w^l}}$ and $s$.
Due to the diversity of the text domain, text with few permutations may share similar semantics. Therefore, we follow the common practice of using the conditioning augmentation method \cite{zhang2017stackgan} to augment the text descriptions. This produces more image-text pairs and thus encourages robustness to small perturbations along the conditioning text manifold. Specifically, we use $F_{ca}$ to represent the conditioning augmentation function and obtain the augmented sentence vector: 
\begin{equation}
{s_{ca}} = {F_{ca}}\left( s \right),
\label{eq:sentenceAug}
\end{equation}
where $s_{ca}\in \mathbb{R}^{D'}$, $D'$ is the dimension after augmentation.

\subsection{GLAM: Global-Local collaborative Attentive Module in Cascaded Image Generators}
We next construct a multi-stage cascaded generator by stacking three image generation networks sequentially. We adopt the basic structure described in \cite{xu2017attngan} due to its good performance in generating realistic images. Mathematically, we use \{$F_0$, $F_1$, ..., $F_{m-1}$\} to denote the $m$ visual feature transformers and \{$G_0$, $G_1$, ..., $G_{m-1}$\} to denote the $m$ image generators. The visual feature $f_i$ and generated image $I_i$ in each stage can be expressed as:
\begin{align}\nonumber
&{f_0} = {F_0}\left( {z,{s_{ca}}} \right), \\ \nonumber
    &{f_i} = {F_i}\left( {{f_{i - 1}},{F_{at{t_i}}}\left( {{f_{i - 1}},w,{s_{ca}}} \right)} \right),i \in \left\{ {1,2, \ldots ,m - 1} \right\}, \\
    &{I_i} = {G_i}\left( {{f_i}} \right),i \in \left\{ {0,1,2, \ldots ,m - 1} \right\},
\label{eq:glam}
\end{align}
where $f_i\in \mathbb{R}^{M_i \times N_i}$ and $I_i\in \mathbb{R}^{q_i \times q_i}$, $z\sim  N(0, 1)$ denotes random noises. $F_{att_i}$ is the proposed global-local collaborative attention model which includes two components ${Att_{i - 1}^w}$ and ${Att_{i - 1}^s}$, $i.e.,$ ${F_{at{t_i}}}\left( {{f_{i - 1}},w,{s_{ca}}} \right) = concat\left( {Att_{i - 1}^w,Att_{i - 1}^s} \right)$.

First, we use the word-level attention model proposed in \cite{xu2017attngan} to generate an attentive word-context feature. It takes the word embedding $w$ and the visual feature $f$ as the input in each stage. The word embedding $w$ is first converted into an underlying common semantic space of visual features by a perception layer $U_{i-1}$ as $U_{i-1}w$. Then, it is multiplied with the visual feature ${f_{i-1}}$ to obtain the attention score. Finally, the attentive word-context feature is obtained by calculating the inner product between the attention score and $U_{i-1}w$: 
\begin{equation}
Att_{i - 1}^w = \sum\limits_{l = 0}^{L - 1} {\left( {{U_{i - 1}}w^l} \right){{\left( {softmax\left( {f_{i - 1}^T\left( {{U_{i - 1}}w^l} \right)} \right)} \right)}^T}},
\label{eq:wordAttention}
\end{equation}
where $U_{i-1}\in \mathbb{R}^{M_{i-1}\times D}$ and $Att_{i-1}^{w} \in \mathbb{R}^{M_{i-1}\times N_{i-1}}$. The attentive word-context feature $Att_{i - 1}^w$ has the exact same dimension as $f_{i-1}$, which is further used for generating the $i^{th}$ visual features $f_i$ by concatenation with $f_{i-1}$.

Then, we propose a sentence-level attention model to enforce a global constraint on the generators during generation. By analogy to the word-level attention model, the augmented sentence vector $s_{ca}$ is first converted into an underlying common semantic space of visual features by a perception layer $V_{i-1}$ as $V_{i-1}s_{ca}$. Then, it is element-wise multiplied with the visual feature ${f_{i-1}}$ to obtain the attention score. Finally, the attentive sentence-context feature is obtained by calculating the element-wise multiplication of the attention score and $V_{i-1}s_{ca}$: 
\begin{equation}
Att_{i - 1}^s = \left( {{V_{i - 1}}{s_{ca}}} \right) \circ \left( {softmax\left( {{f_{i - 1}} \circ \left( {{V_{i - 1}}{s_{ca}}} \right)} \right)} \right),
\label{eq:sentenceAttention}
\end{equation}
where $\circ$ denotes the element-wise multiplication, $V_{i}\in \mathbb{R}^{M_{i}\times D'}$ and $Att_{i-1}^{s} \in \mathbb{R}^{M_{i-1}\times N_{i-1}}$. The attentive sentence-context feature $Att_{i - 1}^s$ is further concatenated with $f_{i-1}$ and $Att_{i - 1}^w$ for generating the $i^{th}$ visual features $f_i$ as depicted in the second equality in Eq.~\eqref{eq:glam}.

\subsection{STREAM: Semantic Text REgeneration and Alignment Module}
As described above, MirrorGAN includes a semantic text regeneration and alignment module (STREAM) to regenerate the text description from the generated image, which semantically aligns with the given text description. Specifically, we employ a widely used encoder-decoder-based image caption framework \cite{karpathy2015deep,vinyals2015show} as the basic STREAM architecture. Note that a more advanced image captioning model can also be used, which is likely to produce better results. However, in a first attempt to validate the proposed idea, we simply exploit the baseline in the current work. 

The image encoder is a convolutional neural network (CNN) \cite{he2016deep} pretrained on ImageNet \cite{deng2009imagenet}, and the decoder is a RNN \cite{hochreiter1997long}. The image $I_{m-1}$ generated by the final stage generator is fed into the CNN encoder and RNN decoder as follows: 
\begin{equation}
\begin{split}
&x_{-1} = CNN(I_{m-1}),   \\
&x_t = W_e T_t, t\in \{0,...L-1\}, \\
&p_{t+1} = RNN(x_t), t\in \{0,...L-1\}, \\
\end{split}
\end{equation}
where $x_{-1}\in \mathbb{R}^{M_{m-1}}$ is a visual feature used as the input at the beginning to inform the RNN about the image content. $W_e\in \mathbb{R}^{M_{m-1} \times D}$ represents a word embedding matrix, which maps word features to the visual feature space. $p_{t+1}$ is a predicted probability distribution over the words. We pre-trained STREAM as it helped MirrorGAN achieve a more stable training process and converge faster, while jointly optimizing STREAM with MirrorGAN is instable and very expensive in terms of time and space. The encoder-decoder structure in \cite{vinyals2015show} and then their parameters keep fixed when training the other modules of MirrorGAN.


\subsection{Objective functions}
Following common practice, we first employ two adversarial losses: a visual realism adversarial loss and a text-image paired semantic consistency adversarial loss, which are defined as follows.

During each stage of training MirrorGAN, the generator $G$ and discriminator $D$ are trained alternately. Specially, the generator $G_{i}$ in the $i^{th}$ stage is trained by minimizing the loss as follows:
\begin{equation}
\begin{array}{c}
{\mathcal{L}_{{G_i}}} =  - \frac{1}{2}{E_{{I_i} \sim {p_{{I_i}}}}}\left[ {\log \left( {{D_i}\left( {{I_i}} \right)} \right)} \right]  \\
- \frac{1}{2}{\mathbb{E}_{{I_i} \sim {p_{{I_i}}}}}\left[ {\log \left( {{D_i}\left( {{I_i},s} \right)} \right)} \right],
\end{array}
\end{equation}
where $I_i$ is a generated image sampled from the distribution $p_{I_i}$ in the $i^{th}$ stage. The first term is the visual realism adversarial loss, which is used to distinguish whether the image is visually real or fake, while the second term is the text-image paired semantic consistency adversarial loss, which is used to determine whether the underlying image and sentence semantics are consistent.

We further propose a CE-based text-semantic reconstruction loss to align the underlying semantics between the redescription of STREAM and the given text description. Mathematically, this loss can be expressed as: 
\begin{equation}
\mathcal{L}_{stream} = -\sum_{t=0}^{L-1}log\:p_{t}(T_t). 
\end{equation}
It is noteworthy that $\mathcal{L}_{stream}$ is also used during STREAM pretraining. When training $G_{i}$, gradients from $\mathcal{L}_{stream}$ are backpropagated to $G_{i}$ through STREAM, whose network weights are kept fixed.

The final objective function of the generator is defined as: 
\begin{equation} \label{captioning}
{\mathcal{L}_G} = \sum\limits_{i = 0}^{m - 1} {{\mathcal{L}_{{G_i}}}}  + \lambda {\mathcal{L}_{stream}}, 
\end{equation}
where $\lambda$ is a loss weight to handle the importance of adversarial loss and the text-semantic reconstruction loss.

The discriminator $D_i$ is trained alternately to avoid being fooled by the generators by distinguishing the inputs as either real or fake. Similar to the generator, the objective of the discriminators consists of a visual realism adversarial loss and a text-image paired semantic consistency adversarial loss. Mathematically, it can be defined as:
\begin{equation}
\begin{array}{c}
 {\mathcal{L}_{{D_i}}} =  - \frac{1}{2}{\mathbb{E}_{I_i^{GT} \sim {p_{I_i^{GT}}}}}\left[ {\log \left( {{D_i}\left( {I_i^{GT}} \right)} \right)} \right]  \\
 - \frac{1}{2}{\mathbb{E}_{{I_i} \sim {p_{{I_i}}}}}\left[ {\log \left(1- {{D_i}\left( {{I_i}} \right)} \right)} \right] \\
  - \frac{1}{2}{\mathbb{E}_{I_i^{GT} \sim {p_{I_i^{GT}}}}}\left[ {\log \left(  {{D_i}\left( {I_i^{GT},s} \right)} \right)} \right]  \\
  - \frac{1}{2}{\mathbb{E}_{{I_i} \sim {p_{{I_i}}}}}\left[ {\log \left( 1-{{D_i}\left( {{I_i},s} \right)} \right)} \right], \\
 \end{array}
\end{equation}
where $I_i^{GT}$ is from the real image distribution $p_{I_i^{GT}}$ in $i^{th}$ stage. 
The final objective function of the discriminator is defined as:
\begin{equation}
{\mathcal{L}_D} = \sum\limits_{i = 0}^{m - 1} {{\mathcal{L}_{{D_i}}}}. 
\end{equation}

\section{Experiments}
In this section, we present extensive experiments that evaluate the proposed model. 
We first compare MirrorGAN with the state-of-the-art T2I methods GAN-INT-CLS \cite{reed2016generative}, GAWWN \cite{reed2016learning}, StackGAN \cite{zhang2017stackgan}, StackGAN++ \cite{zhang2017stackgan++}, PPGN \cite{nguyen2017plug} and AttnGAN \cite{xu2017attngan}. 
Then, we present ablation studies on the key components of MirrorGAN including GLAM and STREAM.

\subsection{Experiment setup}
\subsubsection{Datasets}
We evaluated our model on two commonly used datasets, CUB bird dataset \cite{wah2011caltech} and MS COCO dataset \cite{lin2014microsoft}. The CUB bird dataset contains 8,855 training images and 2,933 test images belonging to 200 categories, each bird image has 10 text descriptions. The COCO dataset contains 82,783 training images and 40,504 validation images, each image has 5 text descriptions. Both datasets were pre-processed using the same pipeline as in \cite{zhang2017stackgan,xu2017attngan}.
\subsubsection{Evaluation metric}
Following common practice \cite{zhang2017stackgan,xu2017attngan}, the Inception Score \cite{salimans2016improved} was used to measure both the objectiveness and diversity of the generated images. Two fine-tuned inception models provided by \cite{zhang2017stackgan} were used to calculate the score.

Then, the R-precision introduced in \cite{xu2017attngan} was used to evaluate the visual-semantic similarity between the generated images and their corresponding text descriptions. For each generated image, its ground truth text description and 99 randomly selected mismatched descriptions from the test set were used to form a text description pool. We then calculated the cosine similarities between the image feature and the text feature of each description in the pool, before counting the average accuracy at three different settings: top-1, top-2, and top-3. The ground truth entry falling into the top-k candidates was treated as correct, otherwise, it was wrong. A higher score represents a higher visual-semantic similarity between the generated images and input text. 

The Inception Score and the R-precision were calculated accordingly as in \cite{zhang2017stackgan,xu2017attngan}.
\subsubsection{Implementation details}
MirrorGAN has three generators in total and GLAM is employed over the last two generators, as shown in Eq.~\eqref{eq:glam}. $64\times 64$, $128\times 128$, $256\times 256$ images are generated progressively. Followed \cite{xu2017attngan}, a pre-trained bi-directional LSTM \cite{schuster1997bidirectional} was used to calculate the semantic embedding from text descriptions. The dimension of the word embedding $D$ was 256. The sentence length $L$ was 18. The dimension $M_{i}$ of the visual embedding was set to 32. The dimension of the visual feature was $N_{i}=q_i \times q_i$, where $q_i$ was 64, 128, and 256 for the three stages. The dimension of augmented sentence embedding $D'$ was set to 100. The loss weight $\lambda$ of the text-semantic reconstruction loss was set to 20.

\subsection{Main results}
In this section, we present both qualitative and quantitative comparisons with other methods to verify the effectiveness of MirrorGAN. First, we compare MirrorGAN with state-of-the-art text-to-image methods \cite{reed2016generative,reed2016learning,zhang2017stackgan,zhang2017stackgan++,nguyen2017plug,xu2017attngan} using the Inception Score and R-precision score on both CUB and COCO datasets. Then, we present subjective visual comparisons between MirrorGAN and the state-of-the-art method AttnGAN \cite{xu2017attngan}. We also present the results of a human study designed to test the authenticity and visual semantic similarity between input text and images generated by MirrorGAN and AttnGAN \cite{xu2017attngan}.

\subsubsection{Quantitative results}\label{InceptionScore}
The Inception Scores of MirrorGAN and other methods are shown in Table \ref{tab:InceptionScore}. MirrorGAN achieved the highest Inception Score on both CUB and COCO datasets. Specifically, compared with the state-of-art method AttnGAN \cite{xu2017attngan}, MirrorGAN improved the Inception Score from 4.36 to 4.56 on CUB and from 25.89 to 26.47 on the more difficult COCO dataset. These results show that MirrorGAN can generate more diverse images of better quality.
\begin{figure*}[tb!]
\centering
\noindent\includegraphics[width=1.9\columnwidth]{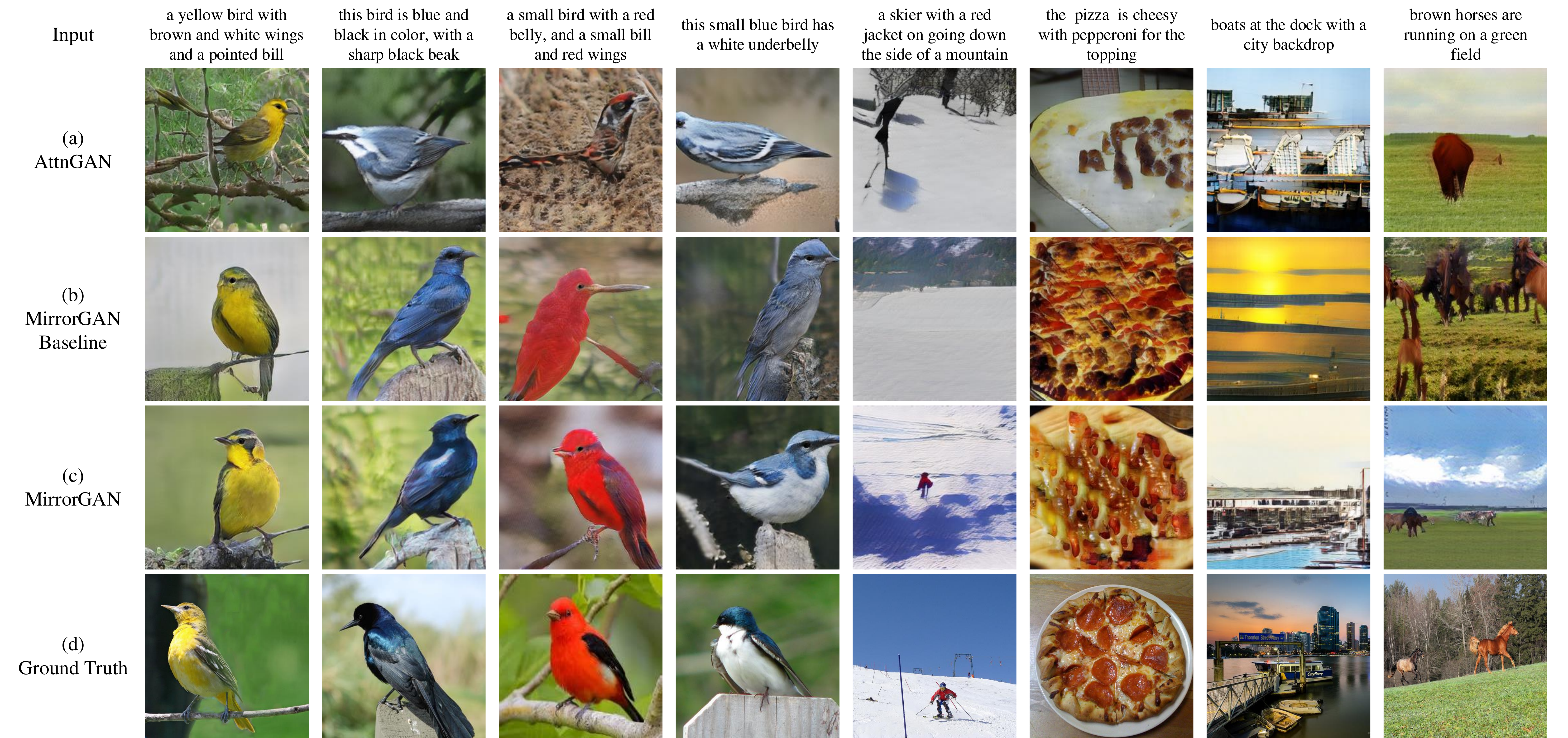}
\protect\caption{Examples of images generated by (a) AttnGAN \cite{xu2017attngan}, (b) MirrorGAN Baseline, and (c) MirrorGAN conditioned on text descriptions from CUB and COCO test sets and (d) the corresponding ground truth.}
\label{fig:Visualization}
\end{figure*}

\begin{table}[htbp]
\centering
\fontsize{7.5}{9}\selectfont
\caption{\label{tab:InceptionScore}Inception Scores of state-of-the-art methods and MirrorGAN on CUB and COCO datasets.}
\begin{tabular}{lcc}
\toprule
\textbf{Model} & \textbf{CUB}& \textbf{COCO}\\
\cmidrule{1-3}
GAN-INT-CLS \cite{reed2016generative}                                                                        & {2.88 $\pm$ 0.04}         & {7.88 $\pm$ 0.07}       \\
GAWWN \cite{reed2016learning}                                                                                     & {3.62 $\pm$ 0.07}        & {-}                                 \\
StackGAN \cite{zhang2017stackgan}                                                                               & {3.70 $\pm$ 0.04}       & {8.45 $\pm$ 0.03}      \\
StackGAN++ \cite{zhang2017stackgan++}                                                                      & {3.82 $\pm$ 0.06}       & {-}                                 \\
PPGN \cite{nguyen2017plug}                                                                                             & {-}                                  & {9.58 $\pm$ 0.21}       \\
AttnGAN \cite{xu2017attngan}                                                                                            & {4.36 $\pm$ 0.03}        & {25.89 $\pm$ 0.47}     \\
\midrule
MirrorGAN                                                                                                                        &  \textbf{4.56 $\pm$ 0.05}          &  \textbf{26.47 $\pm$ 0.41}     \\
\bottomrule
\end{tabular}
\end{table}

The R-precision scores of AttnGAN \cite{xu2017attngan} and MirrorGAN on CUB and COCO datasets are listed in Table \ref{tab:R-precision}. MirrorGAN consistently outperformed AttnGAN \cite{xu2017attngan} at all settings by a large margin, demonstrating the superiority of the proposed text-to-image-to-text framework and the global-local collaborative attentive module, since MirrorGAN generated high-quality images with semantics consistent with the input text descriptions.

\begin{table}
\centering
\fontsize{7.5}{9}\selectfont
\caption{\label{tab:R-precision} R-precision [\%] of the state-of-the-art AttnGAN \cite{xu2017attngan} and MirrorGAN on CUB and COCO datasets.}
\begin{tabular}{>{}l*{6}{c}}\toprule
{\bfseries Dataset} & \multicolumn{3}{c}{\bfseries CUB}
                                  & \multicolumn{3}{c} {\bfseries COCO}
                                   \\\cmidrule(lr){2-4}\cmidrule(lr){5-7}
{top-k}             & k=1 & k=2   & k=3    & k=1& k=2 & k=3\\ \midrule
AttnGAN \cite{xu2017attngan}                & {53.31}        & {54.11}   & {54.36}    & {72.13}        & {73.21}   & {76.53}  \\
MirrorGAN                    & \textbf{57.67}&  \textbf{58.52}  & \textbf{60.42}   & \textbf{74.52}   & \textbf{76.87}    & \textbf{80.21} \\
    \bottomrule
\end{tabular}
\end{table}

\subsubsection{Qualitative results} \label{Humanstudy}
\textbf{Subjective visual comparisons: }Subjective visual comparisons between AttnGAN \cite{xu2017attngan}, MirrorGAN Baseline, and MirrorGAN are presented in Figure~\ref{fig:Visualization}. MirrorGAN Baseline refers to the model using only word-level attention for each generator in the MirrorGAN framework.

It can be seen that the image details generated by AttnGAN are lost, colors are inconsistent with the text descriptions ($3^{rd}$ and $4^{th}$ column), and the shape looks strange ($2^{nd}$, $3^{rd}$, $5^{th}$ and $8^{th}$ column) for some hard examples. Furthermore, the skier is missing in the $5^{th}$ column. MirrorGAN Baseline achieved better results with more details and consistent colors and shapes compared to AttnGAN. For example, the wings are vivid in the $1^{st}$ and $2^{nd}$ columns, demonstrating the superiority of MirrorGAN and that it takes advantage of the dual regularization by redescription, $i.e.,$ a semantically consistent image should be generated if it can be redescribed correctly. By comparing MirrorGAN with MirrorGAN Baseline, we can see that GLAM contributes to producing fine-grained images with more details and better semantic consistency. For example, the color of the underbelly of the bird in the $4^{th}$ column was corrected to white, and the skier with a red jacket was recovered. The boats and city backdrop in the $7^{th}$ column and the horses on the green field in the $8^{th}$ column look real at first glance. Generally, content in the CUB dataset is less diverse than in COCO dataset. Therefore, it is easier to generate visually realistic and semantically consistent results on CUB. These results confirm the impact of GLAM, which uses global and local attention collaboratively.

\begin{figure}[tb!]
\centering
\noindent\includegraphics[width=0.8\columnwidth]{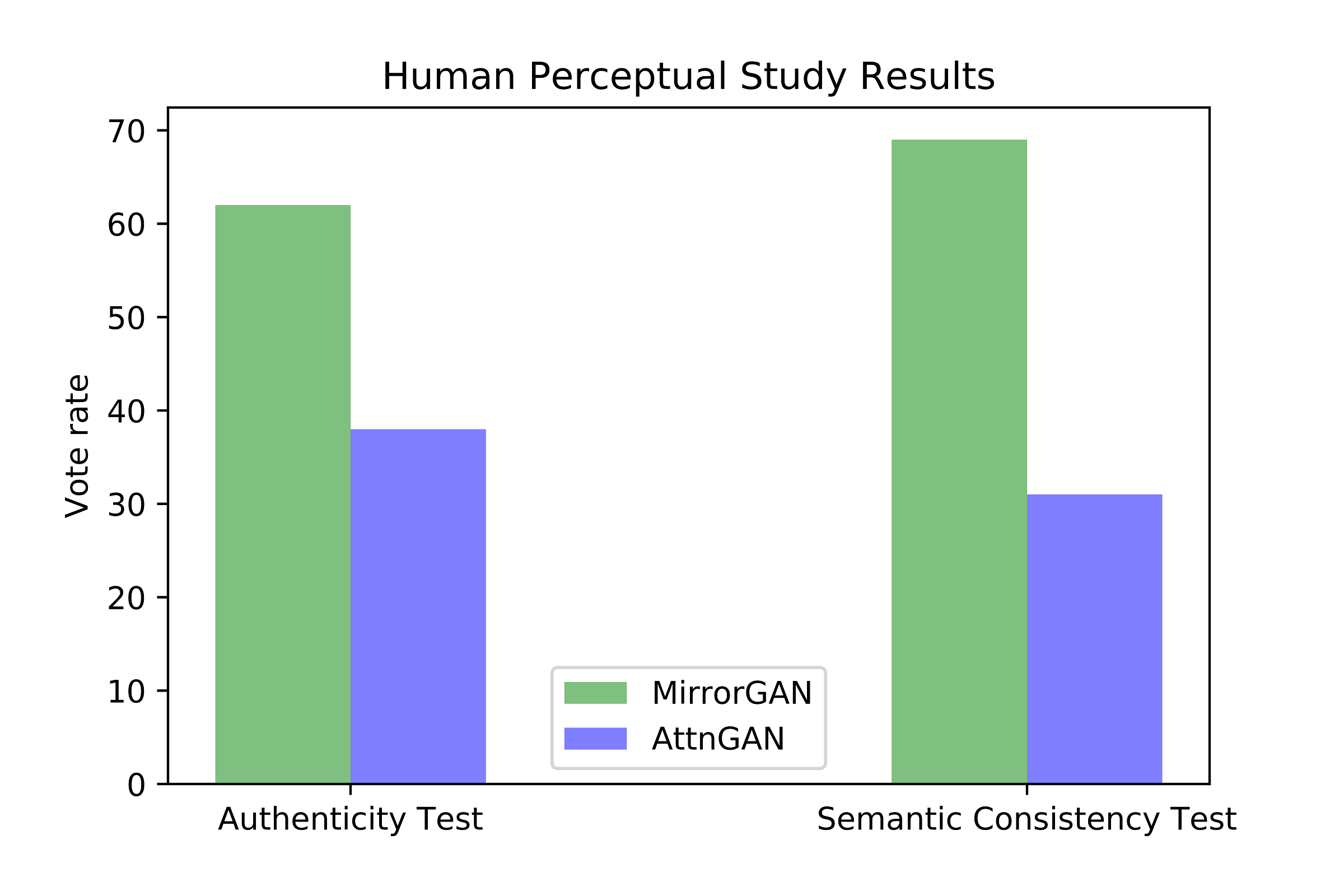}
\protect\caption{Results of Human perceptual test. A higher value of the Authenticity Test means more convincing images. A higher value of the Semantic Consistency Test means a closer semantics between input text and generated images.}
\label{fig:Human}
\end{figure}

\textbf{Human perceptual test: }To compare the visual realism and semantic consistency of the images generated by AttnGAN and MirrorGAN, we next performed a human perceptual test on the CUB test dataset. We recruited 100 volunteers with different professional backgrounds to conduct two tests: the Image Authenticity Test and the Semantic Consistency Test. The Image Authenticity Test aimed to compare the authenticity of the images generated using different methods. Participants were presented with 100 groups of images consecutively. Each group had 2 images arranged in random order from AttnGAN and MirrorGAN given the same text description. Participants were given unlimited time to select the more convincing images. The Semantic Consistency Test aimed to compare the semantic consistency of the images generated using different methods. Each group had 3 images corresponding to the ground truth image and two images arranged at random from AttnGAN and MirrorGAN. The participants were asked to select the images that were more semantically consistent with the ground truth. Note that we used ground truth images instead of the text descriptions since it is easier to compare the semantics between images.

After the participants finished the experiment, we counted the votes for each method in the two scenarios. The results are shown in Figure~\ref{fig:Human}.  It can be seen that the images from MirrorGAN were preferred over ones from AttnGAN. MirrorGAN outperformed AttnGAN with respect to authenticity, MirrorGAN was even more effective in terms of semantic consistency. These results demonstrate the superiority of MirrorGAN for generating visually realistic and semantically consistent images.

\subsection{Ablation studies}
\begin{figure*}[tb!]
\centering
\noindent\includegraphics[width=1.9\columnwidth]{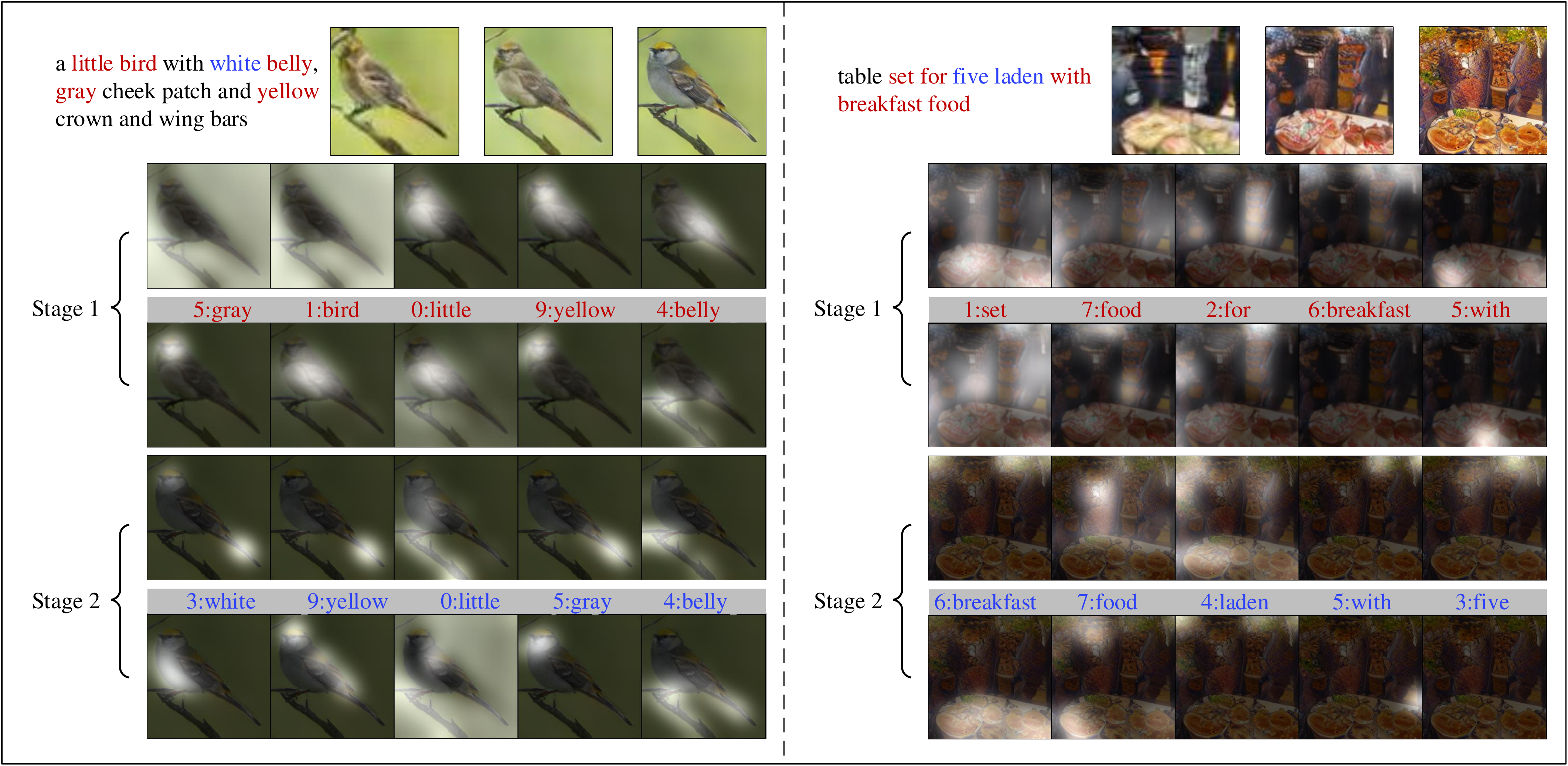}
\protect\caption{Attention visualization on the CUB and the COCO test sets. The first row shows the output $64 \times 64$ images generated by $G_{0}$, $128\times 128$ images generated by $G_{1}$ and $256\times 256$ images generated by $G_{2}$. And the following rows show the Global-Local attention generated in stage 1 and 2. Please refer to the supplementary material for more examples.}
\label{fig:Attention}
\end{figure*}

\begin{table}
\centering
\fontsize{7.5}{9}\selectfont
\caption{\label{tab:ablation}Inception Score and R-precision results of MirrorGAN with different weight settings.}
\begin{tabular}{>{}l*{4}{c}}\toprule
\multirow{2}{*}{\bfseries Evaluation Metric} & \multicolumn{2}{c}{\bfseries Inception Score}
                                                           & \multicolumn{2}{c} {\bfseries R-precision (top-1)}
                                                             \\\cmidrule(lr){2-3}\cmidrule(lr){4-5}
                       & \textbf{CUB} & \textbf{COCO}       & \textbf{CUB}& \textbf{COCO} \\ \midrule
MirrorGAN w/o GA, $\lambda$=0                          & {3.91$\pm$ .09}        & {19.01$\pm$ .42}                 & {39.09}        & {50.69}        \\
MirrorGAN w/o GA, $\lambda$=20            & {4.47$\pm$ .07}        & {25.99$\pm$ .41} & {55.67}            & {73.28}     \\
MirrorGAN, $\lambda$=5                                &{4.01$\pm$ .06}        & {21.85$\pm$ .43}       & {32.07}        & {52.55}      \\
MirrorGAN, $\lambda$=10                              & {4.30$\pm$ .07}        & {24.11$\pm$ .31}       & {43.21}        & {63.40}     \\
MirrorGAN, $\lambda$=20                              &  \textbf{4.54 $\pm$ .17} &  \textbf{26.47$\pm$ .41}  & \textbf{57.67}       & \textbf{74.52}   \\
    \bottomrule
\end{tabular}
\end{table}

\textbf{Ablation studies on MirrorGAN components}:
We next conducted ablation studies on the proposed model and its variants. To validate the effectiveness of STREAM and GLAM, we conducted several comparative experiments by excluding/including these components in MirrorGAN. The results are listed in Table \ref{tab:ablation}. 

First, the hyper-parameter $\lambda$ is important. A larger $\lambda$ led to higher Inception Scores and R-precision on both datasets. On the CUB dataset, 
when $\lambda$ increased from 5 to 20, the Inception Score increased from 4.01 to 4.54 and R-precision increased from 32.07$\%$ to 57.67$\%$. 
On the COCO dataset, the Inception Score increased from 21.85 to 26.21 and R-precision increased from 52.55$\%$ to 74.52$\%$. We set $\lambda$ to 20 as default.

MirrorGAN without STREAM ($\lambda = 0$) and global attention (GA) achieved better results than StackGAN++ \cite{zhang2017stackgan++} and PPGN \cite{nguyen2017plug}. Integrating STREAM into MirrorGAN led to further significant performance gains. The Inception Score increased from 3.91 to 4.47 and from 19.01 to 25.99 on CUB and COCO, respectively, and R-precision showed the same trend. Note that MirrorGAN without GA already outperformed the state-of-the-art AttnGAN (Table~\ref{tab:InceptionScore}) which also used the word-level attention. These results indicate that STREAM is more effective in helping the generators achieve better performance. This attributes to the introduction of a more strict semantic alignment between generated images and input text, which is provided by STREAM. Specifically, STREAM forces the generated images to be redescribed as the input text sequentially, which potentially prevents possible mismatched visual-text concept. Moreover, MirrorGAN integration with GLAM further improved the Inception Score and R-precision to achieve new state-of-the-art performance. These results show that the global and local attention in GLAM collaboratively help the generator to generate visually realistic and semantically consistent results by telling it where to focus on.

\textbf{Visual inspection on the cascaded generators}: 
To better understand the cascaded generation process of MirrorGAN, we visualized both the intermediate images and the attention maps in each stage (Figure~\ref{fig:Attention}). In the first stage, low-resolution images were generated with primitive shape and color but lacking details. With guidance from GLAM in the following stages, MirrorGAN generated images by focusing on the most relevant and important areas. Consequently, the quality of the generated images progressively improved, $e.g.,$ the colors and details of the wings and crown. The top-5 global and local attention maps in each stage are shown below the images. It can be seen that: 1) the global attention concentrated more on the global context in the earlier stage and then the context around specific regions in later stages, 2) the local attention helped the generator synthesize images with fine-grained details by guiding it to focus on the most relevant words, and 3) the global attention is complementary to the local attention, they collaboratively contributed to the progressively improved generation.

\begin{figure}[tb!]
\centering
\noindent\includegraphics[width=1.0\columnwidth]{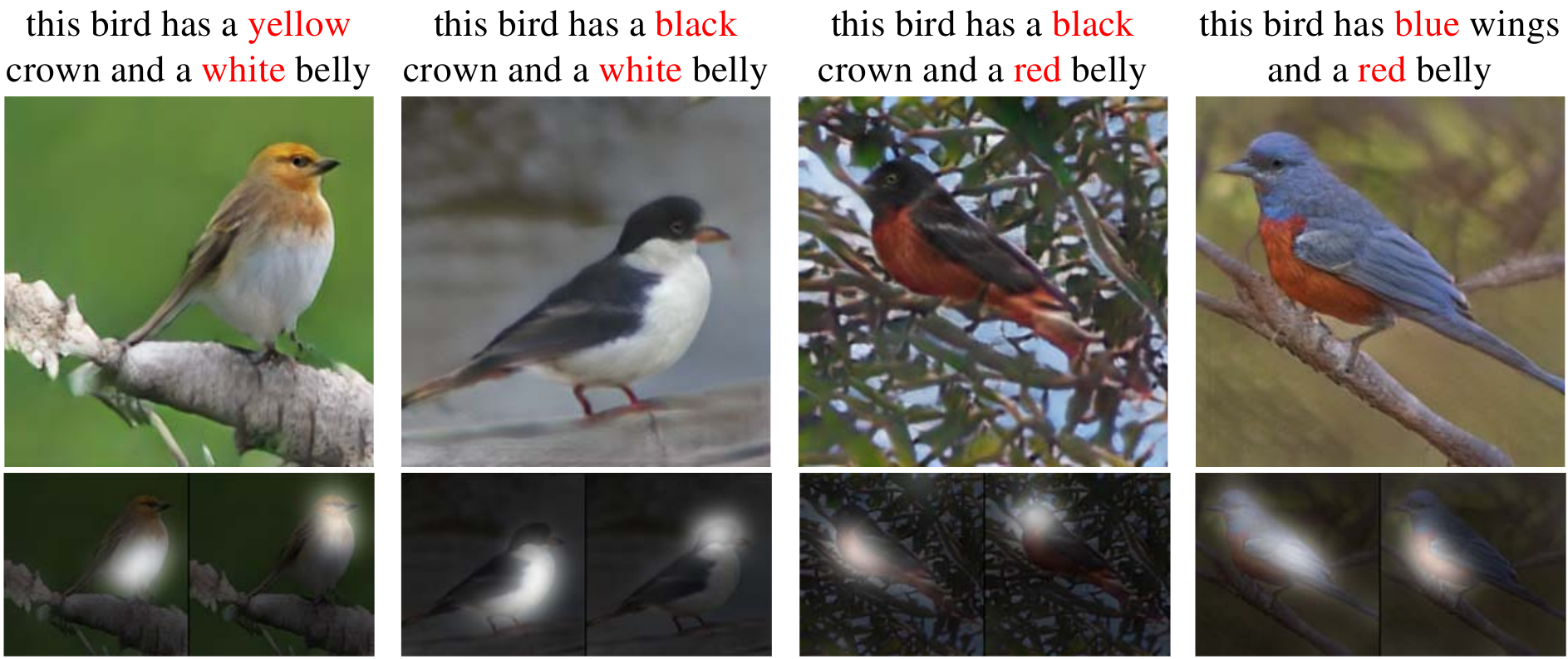}
\protect\caption{Images generated by MirrorGAN by modifying the text descriptions by a single word and the corresponding top-2 attention maps in the last stage.}
\label{fig:colorVariants}
\end{figure}

In addition, we also present the images generated by MirrorGAN by modifying the text descriptions by a single word (Figure~\ref{fig:colorVariants}). MirrorGAN captured subtle semantic differences in the text descriptions.

\subsection{Limitation and discussion}
Although our proposed MirrorGAN shows superiority in generating visually realistic and semantically consistent images, some limitations must be taken into consideration in future studies. First, STREAM and other MirrorGAN modules are not jointly optimized with complete end-to-end training due to limited computational resources. Second, we only utilize a basic method for text embedding in STEM and image captioning in STREAM, which could be further improved, for example, by using the recently proposed BERT model \cite{devlin2018bert} and state-of-the-art image captioning models \cite{anderson2017bottom, Chen_2018_CVPR}. Third, although MirrorGAN is initially designed for the T2I generation by aligning cross-media semantics, we believe that its complementarity to the state-of-the-art CycleGAN can be further exploited to enhance model capacity for jointly modeling cross-media content.

\section{Conclusions}
In this paper, we address the challenging T2I generation problem by proposing a novel global-local attentive and semantic-preserving text-to-image-to-text framework called MirrorGAN. MirrorGAN successfully exploits the idea of learning text-to-image generation by redescription. STEM generates word- and sentence-level embeddings. GLAM has a cascaded architecture for generating target images from coarse to fine scales, leveraging both local word attention and global sentence attention to progressively enhance the diversity and semantic consistency of the generated images. STREAM further supervises the generators by regenerating the text description from the generated image, which semantically aligns with the given text description. We show that MirrorGAN achieves new state-of-the-art performance on two benchmark datasets.

\blfootnote{{\bf{Acknowledgements:}} This work is supported in part by Chinsese National Double First-rate Project about digital protection of cultural relics in Grotto Temple and equipment upgrading of the Chinese National Cultural Heritage Administration scientific research institutes, the National Natural Science Foundation of China Project 61806062, and the Australian Research Council Projects FL-170100117, DP-180103424, and IH-180100002.}

{\small
\bibliographystyle{ieee}
\bibliography{cvpr2019}
}

\end{document}